\documentclass{article}
\usepackage{spconf,amsmath,mathtools,caption,multirow,graphicx,enumerate,xfrac,array}

\title{Multi-Glimpse LSTM \\ with Color-Depth Feature Fusion for Human Detection}
\name{Hengduo Li $^{1}$, Jun Liu $^{1}_{*}$, Guyue Zhang $^{1}$, Yuan Gao $^{1}$, Yirui Wu $^{2}$\thanks{${*}$ Corresponding author}}
\address{$^{1}$ Fudan University, Shanghai, China\\
$^{2}$ Hohai University, Nanjing, China\\
\{lihd14, ljun, guyuezhang13, gaoyuan14\}@fudan.edu.cn,  wuyirui@hhu.edu.cn
}

\begin{document}
%\ninept
%
\maketitle
\begin{abstract}
With the development of depth cameras such as Kinect and Intel Realsense, RGB-D based human detection receives continuous research attention due to its usage in a variety of applications. In this paper, we propose a new Multi-Glimpse LSTM (MG-LSTM) network, in which multi-scale contextual information is sequentially integrated to promote the human detection performance. Furthermore, we propose a feature fusion strategy based on our MG-LSTM network to better incorporate the RGB and depth information. To the best of our knowledge, this is the first attempt to utilize LSTM structure for RGB-D based human detection. Our method achieves superior performance on two publicly available datasets.
\end{abstract}
\begin{keywords}
Human detection, RGB-D, LSTM, Feature fusion 
\end{keywords}
\section{Introduction}
\label{sec:intro}

Human detection has been a hot research area due to its wide usage in video surveillance, self-driving vehicles, human-machine interaction, and robotics. With the development of depth cameras like Kinect and Intel Realsense, various vision-based applications are boosted with the depth information acquired by the devices which are more robust against illumination and texture variations. Among these applications, RGB-D based human detection receives continuous research attention recently.

Recent years have seen a considerable amount of work to solve the RGB-D based human detection problem. Spinello and Arras \cite{HOD} takes inspiration from Histogram of Oriented Gradients (HOG) \cite{HOG} and proposes Histogram of Oriented Depths (HOD) to detect people in dense depth data. A reversible jump Markov chain Monte Carlo (RJ-MCMC) particle filtering method was proposed for human detection and tracking on both fixed and moving color-depth cameras \cite{wongui}. Bagautdinov et al. \cite{cvpr15} proposed a generative model to compute the probabilities of presence of potentially occluding pedestrians from a single depth map provided by RGB-D sensors.

Neural networks have shown their strong capability in a variety of fields, such as object recognition \cite{alexnet}\cite{vgg}\cite{resnet}\cite{final_add1}, activity recognition \cite{ActionRecog1}\cite{ActionRecog2}\cite{final_add2}\cite{final_add3}, semantic segmentation \cite{segmentation1}\cite{segmentation2}, and RGB based human detection \cite{rcnn}. Very recently, Xue et al. \cite{HongyangXue} also explored to apply neural networks for RGB-D based human detection and tracking. A deep CNN was used in their method to identify generated proposals. However, they did not consider the utilization of multi-scale multi-part contextual color-depth information, which is often important for reliable human detection. Based on an observation, when detecting and identifying a target, humans tend to catch a wide-range glimpse to get overview knowledge first, then shrink the area of focus gradually until eyes focus exactly on discriminative parts of the target. Owing to the effective utilization of the contextual information among the multiple glimpses, negative factors like occlusion could be more easily handled by human. Therefore, we propose to explicitly utilize the contextual multi-scale multi-part information for RGB-D based human detection.

Long short-term memory(LSTM) is a powerful neural network structure which is able to model the dynamics and contextual dependences in sequential information. Consequently, in this paper, we propose a Multi-Glimpse LSTM (MG-LSTM) to model the contextual multi-scale color depth information. Besides, to effectively fuse the two modalities of RGB and depth, we further propose a novel fusion strategy for LSTM. In our fusion framework, two bypass LSTM chains take the multi-scale color depth information respectively and fuse it at the main LSTM chain to generate the prediction. This structure can fuse the data flow of RGB and depth more effectively whilst allows the two bypass chains to remain relatively independent. To the best of our knowledge, this is the first attempt to utilize LSTM network for RGB-D based human detection.

The main contributions of this paper are as follows: 

\begin{enumerate}[(1)]
\item We propose a long-short term memory classification model, Multi-Glimpse LSTM (MG-LSTM), for RGB-D human detection. This model takes context features of multi-scale RGB-D data corresponding to each pre-generated proposal as a sequence of input for classification.
\item We propose a fusion strategy of LSTM for RGB-D based human detection composed of two bypass LSTM chains receiving extracted features as input and a main prediction-making LSTM chain.
\end{enumerate}

\section{Multi-Glimplse LSTM with \\Color-Depth Feature Fusion}
\label{sec:method}

Humans tend to utilize multi-scale visual information contextually when detecting targets. Motivated by this observation, we propose our method which includes three stages: proposal generation, multi-scale multi-part feature extracting and classification. Potential pixels of human head-top are localized as proposals at the first stage. A set of multi-scale RGB-D images of each proposal are clipped and forwarded to pre-trained convolutional neural networks to extract fixed-length feature vectors. Finally, the MG-LSTM network with color-depth feature fusion takes extracted features as input for the binary classification.

\begin{figure}[t]
\begin{minipage}[b]{1.0\linewidth}
  \centering
  \centerline{\includegraphics[width=8.1cm]{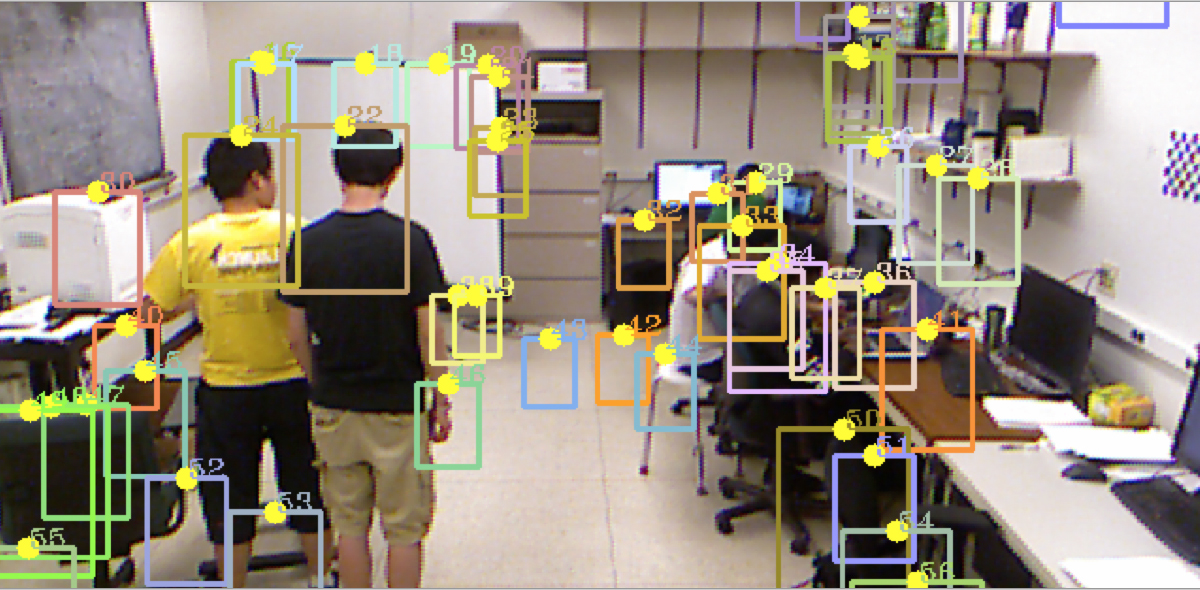}}
%  \vspace{2.0cm}
  \centerline{(a) Example of proposal generation}\medskip
\end{minipage}
\begin{minipage}[b]{.30\linewidth}
  \centering
  \centerline{\includegraphics[width=2.35cm]{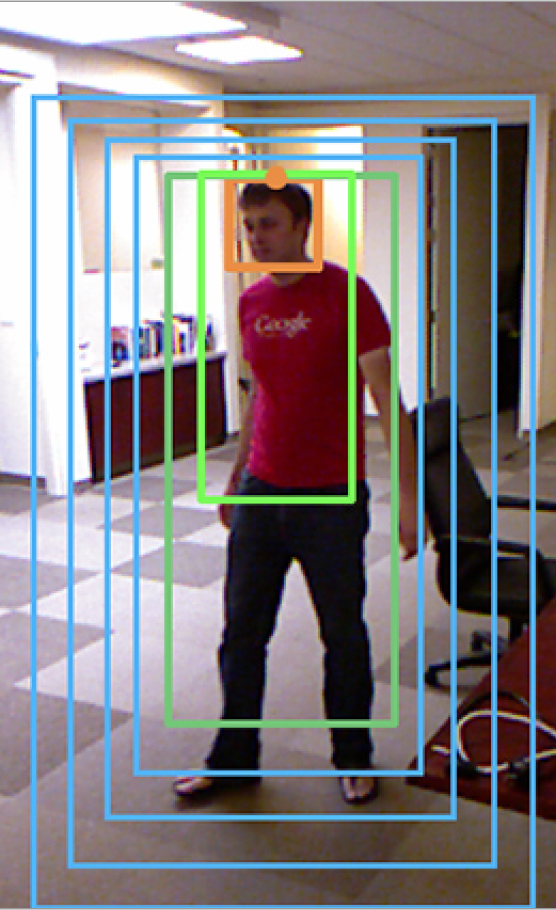}}
%  \vspace{1.5cm}
  \centerline{(b)}\medskip
\end{minipage}
\hfill
\begin{minipage}[b]{0.30\linewidth}
  \centering
  \centerline{\includegraphics[width=2.7cm]{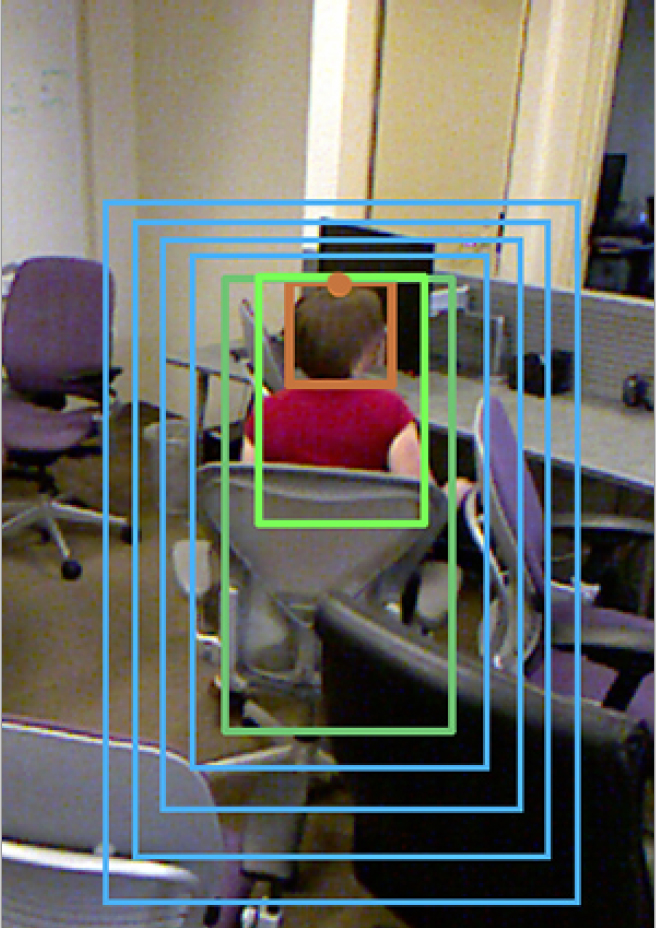}}
%  \vspace{1.5cm}
  \centerline{(c)}\medskip
\end{minipage}
\hfill
\begin{minipage}[b]{0.30\linewidth}
  \centering
  \centerline{\includegraphics[width=2.28cm]{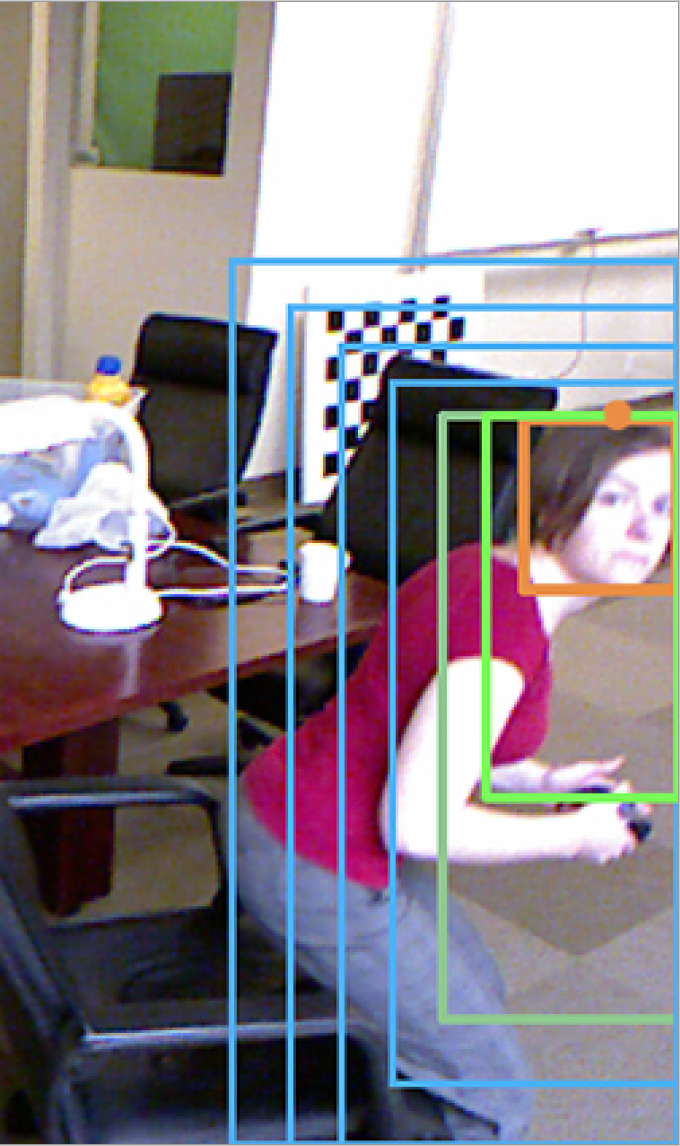}}
%  \vspace{1.5cm}
  \centerline{(d)}\medskip
\end{minipage}
\caption{{\bf Examples of proposal generation (a) and multi-scale images clipping (b-d). (displayed in color domain)} Side of clipping windows stops expansion when reaching borders of original image as shown in (d).}
\label{fig:samples}
\end{figure}

\subsection{Proposal Generation}
\label{ssec:progeneration}

For RGB based human detection, various proposal generation methods are adopted including selective search \cite{selectivesearch}, multi-scale combinatorial grouping \cite{mcg}, objectness \cite{objectness}, etc. In RGB-D based human detection, additional depth information offers opportunity for faster and more reliable proposal generation. Recently, Liu et al \cite{jvci} proposed an ultra-fast proposal generation method, called plausible candidate retriever, for proposal generation in depth image. In this method, every pixel within the depth image is evaluated to judge whether it's a possible location of a human head-top. Plausible head-top pixels are then collected as proposals (candidate human locations). These proposals (candidate human locations) are used for the subsequent human detection procedures.

Compared to other RGB map based approaches, this proposal generation technique proposes much fewer proposals for each image, thus can significantly reduce the time consumption of proposal generation and subsequent processes. According to our experimental results, only around 50 proposals are generated for each image with an extremely small miss rate (0.03). The generated proposals are much fewer than those of Regions with Convolutional Neural Network(R-CNN, around 2,000 region proposals) \cite{rcnn}. Besides, the speed of this process reaches as high as 500 fps. Thus, we use this method for proposal generation. Fig. \ref{fig:samples}(a) shows a typical result of this method. Readers are referred to \cite{jvci} for more details of the proposal generation method.

\subsection{Multi-Glimplse LSTM}
\label{ssec:mglstm}

\begin{figure*}[h]
\begin{minipage}[b]{1.0\linewidth}
  \centering
  \centerline{\includegraphics[width=15cm]{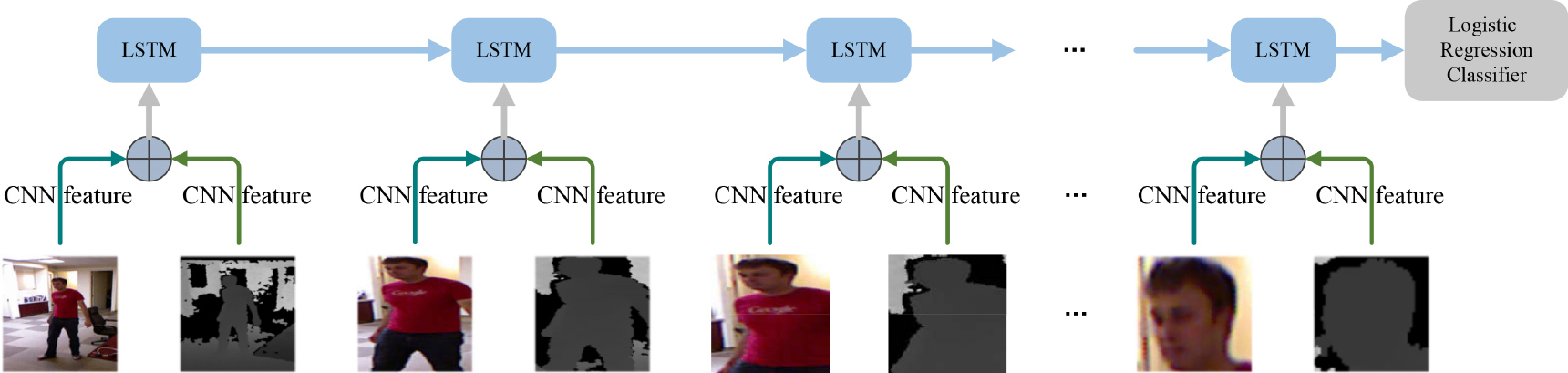}}
%  \vspace{2.0cm}
\end{minipage}
\caption{{\bf Multi-Glimpse LSTM with Color Depth Feature Concatenation.} $\bigoplus$ represents feature concatenation.}
\label{fig:network1}
\end{figure*}

\begin{figure*}[h]
\begin{minipage}[b]{1.0\linewidth}
  \centering
  \centerline{\includegraphics[width=15cm]{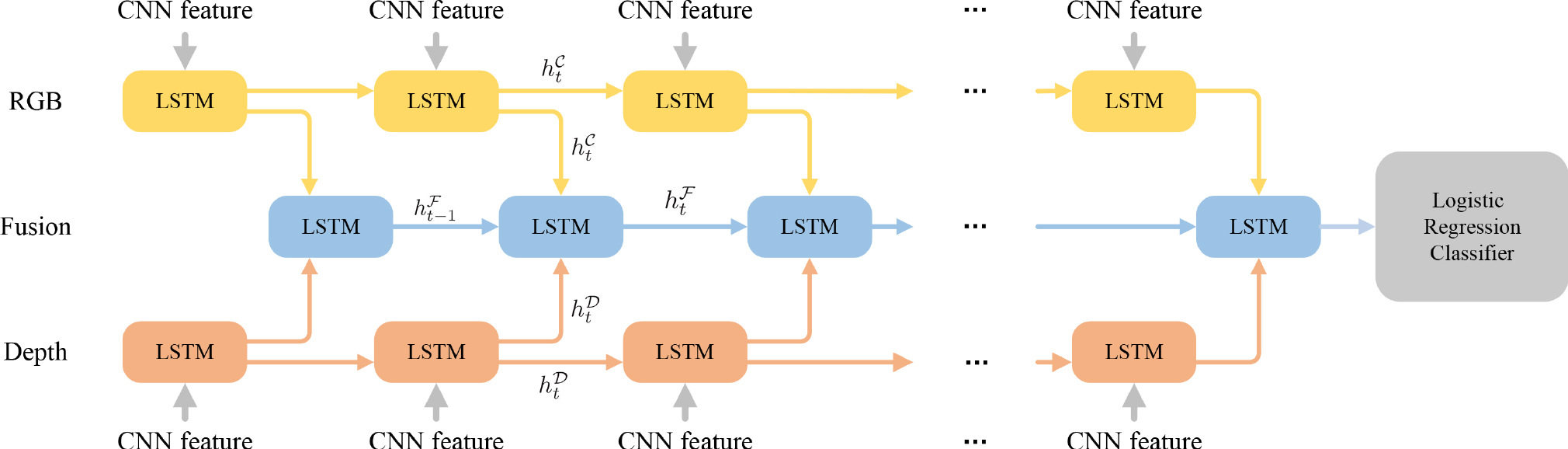}}
\end{minipage}
\caption{{\bf Multi-Glimpse LSTM with Color Depth Feature Fusion Strategy. }}
\label{fig:network2}
\end{figure*}

Human uses contextual multi-scale information when detecting and identifying a target. Similarly, our method utilizes clipped multi-scale multi-part color depth images. As shown in Fig. \ref{fig:samples}(b), for each plausible head-top pixel (candidate human location), a set of multi-scale images are clipped from both color and depth maps. The set of images includes a head-scale, an upperbody-scale, a body-scale and several larger ones. Clipping sizes of head and body are obtained with their real-world sizes, depth value of the plausible head-top pixel and depth camera's intrinsic parameters \cite{jvci}. Size of $n$th peripheral image is calculated as: $S_{n}$ = $S_{b}$ * $(1 + 0.3n)$ in which $S_{b}$ denotes the size of body scale image. 

Arranged contextually in large-to-small order, clipped images are further forwarded to pre-trained CNN networks for extracting discriminative features. For color images, the VGG-19 model \cite{vgg} pre-trained on Imagenet is utilized. For depth images, we use the depth based CNN model trained by Eitel et al. \cite{depthextract}, which performs well on object recognition. A sequence of 4,096 dimensional feature vectors is obtained.

LSTM network, with the ability of modeling the contextual dependencies in a sequence of information, is ideal for modeling this sequence of features. Therefore, we propose the Multi-Glimpse LSTM network (MG-LSTM) which takes the sequence of contextual color depth feature vectors as input to make classification. (Fig. \ref{fig:network1}) The logistic regression classifier completes the binary classification at the last step. The LSTM transition equations are as follows: 
\begin{gather}
\begin{array}{c}
\begin{pmatrix}
i_{t}\\
f_{t}\\
o_{t}\\
u_{t}\\
\end{pmatrix}
\end{array}
=
\begin{array}{c}
\begin{pmatrix}
\sigma\\
\sigma\\
\sigma\\
\tanh\\
\end{pmatrix}
\end{array}
\begin{array}{c}
\begin{pmatrix}
M
\begin{array}{c}
\begin{pmatrix}
x_{t}\\
h_{t-1}\\
\end{pmatrix}
\end{array}
\end{pmatrix}
\end{array}
\\
c_{t} = i_{t} \odot u_{t} + f_{t} \odot c_{t-1} \\
h_{t} = o_{t} \odot \tanh(c_{t})
\end{gather}
where an input gate $i_{t}$, a forget gate $f_{t}$, an output gate $o_{t}$ and formulated input $u_{t}$ are included. $t$ denotes the glimpse step. $\sigma$ is the sigmoid function. $M$ is the affine transformation composed by model parameters. $x$ denotes input of concatenated color depth feature vectors. $h_{t}$ and $c_{t}$ denote the output state and cell state. $\odot$ indicates element-wise production in the formula. 

When modeling the sequence of contextual multi-scale multi-part information, the network gains an overview understanding from the large clipped images firstly, then refinements are added step by step based on the sequential input. Since the multi-scale images within a set are sequential and tightly correlated, a better understanding of the plausible target can be obtained through modeling. Such comprehensive utilization of the contextual information is obviously effective for this binary classification task. Besides, as small-scale images contain human head and upper body which are less deformable whilst large-scale images are more probable to contain intact information, negative impact from occlusion and irregular human poses (Fig. \ref{fig:samples}(c)(d)) can be reduced.

\subsection{Color Depth Feature Fusion}
\label{ssec:fusion}

Color and depth are two different while correlating modalities. In Fig. \ref{fig:network1}, we simply concatenate the two types of CNN features as the input of each step of LSTM. In order to use these two modalities more effectively, in this section, we propose a fusion strategy, in which we merge three chains of LSTM network to achieve color-depth feature fusion. Structure of the fusion network is demonstrated in Fig. \ref{fig:network2}. Within each step, two bypass LSTM chains take color depth feature vectors respectively and fuse them into the main LSTM chain. The logistic regression classifier connected to the last step of the main chain makes binary classification finally.

In our fusion scheme, the gates of the two bypass chains are formulated in the same way of equation (1-3), yet the gates of the main chain are calculated as follows: 
% the following is formula of MG-LSTM
\begin{gather}
\begin{array}{c}
\begin{pmatrix}
i^{\mathcal{F}}_{t}\\
f^{\mathcal{F}}_{t}\\
o^{\mathcal{F}}_{t}\\
u^{\mathcal{F}}_{t}\\
\end{pmatrix}
\end{array}
=
\begin{array}{c}
\begin{pmatrix}
\sigma\\
\sigma\\
\sigma\\
\tanh\\
\end{pmatrix}
\end{array}
\begin{array}{c}
\begin{pmatrix}
M^{\mathcal{F}}
\begin{array}{c}
\begin{pmatrix}
h^{\mathcal{C}}_{t}\\
h^{\mathcal{D}}_{t}\\
h^{\mathcal{F}}_{t-1}\\
\end{pmatrix}
\end{array}
\end{pmatrix}
\end{array}
\end{gather}
where the subscript $\mathcal{C}$, $\mathcal{D}$, $\mathcal{F}$, denote the Color chain, Depth chain and main Fusion chain respectively.

By applying this color-depth feature fusion strategy, color and depth features can be fused effectively through the data flow compared to simple concatenation. Also, the existence of three chains enables the two modalities remain relatively independent instead of being totally interweaved.

\section{Experiment}
\label{sec:exp}

\begin{figure}[t]

\begin{minipage}[b]{0.48\linewidth}
  \centering
  \centerline{\includegraphics[width=4.15cm]{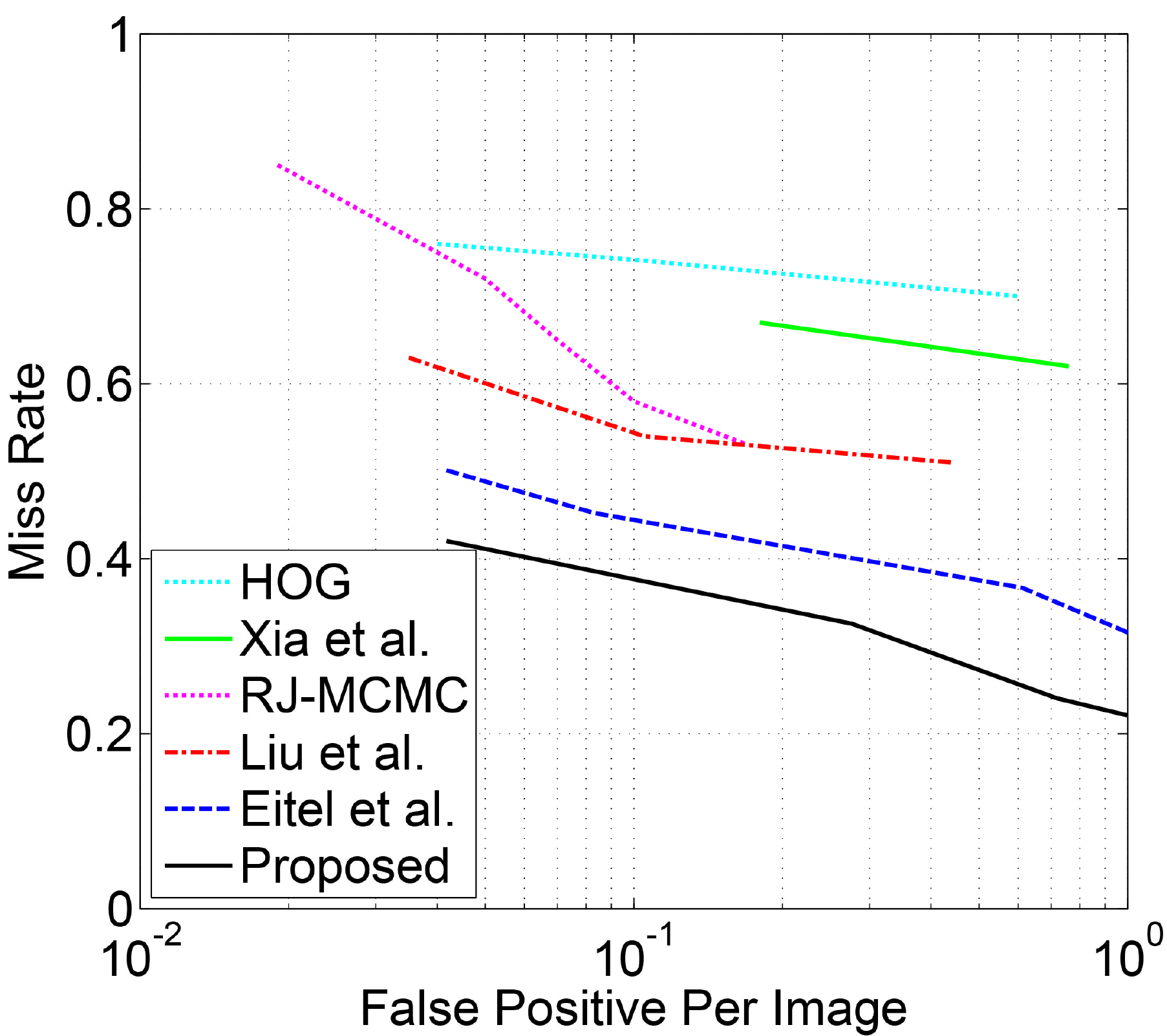}}
%  \vspace{2.0cm}
  \centerline{(a)}\medskip
\end{minipage}
\begin{minipage}[b]{0.51\linewidth}
  \centering
  \centerline{\includegraphics[width=4.2cm]{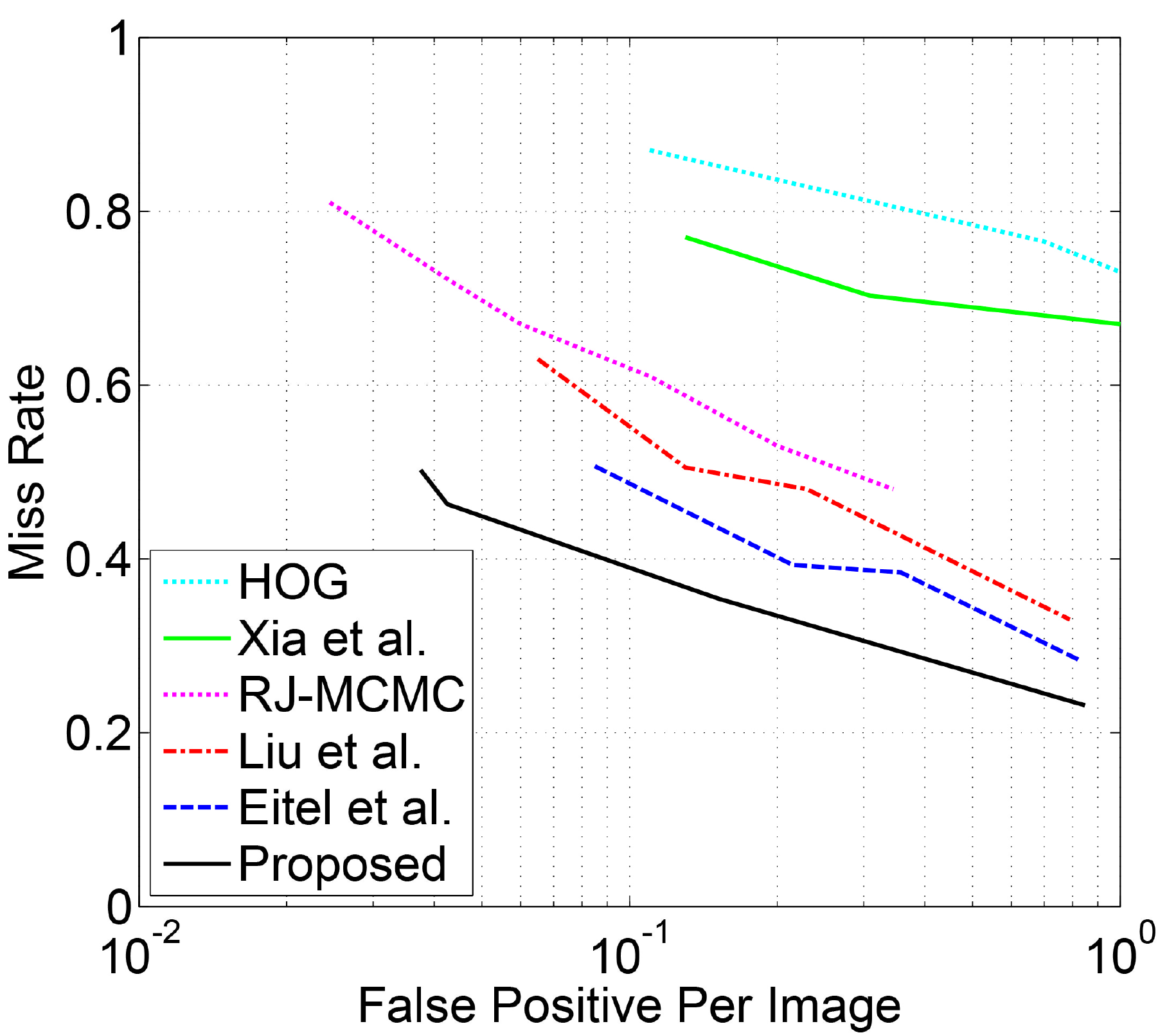}}
%  \vspace{1.5cm}
  \centerline{(b)}\medskip
\end{minipage}
\hfill
\begin{minipage}[b]{0.48\linewidth}
  \centering
  \centerline{\includegraphics[width=4.15cm]{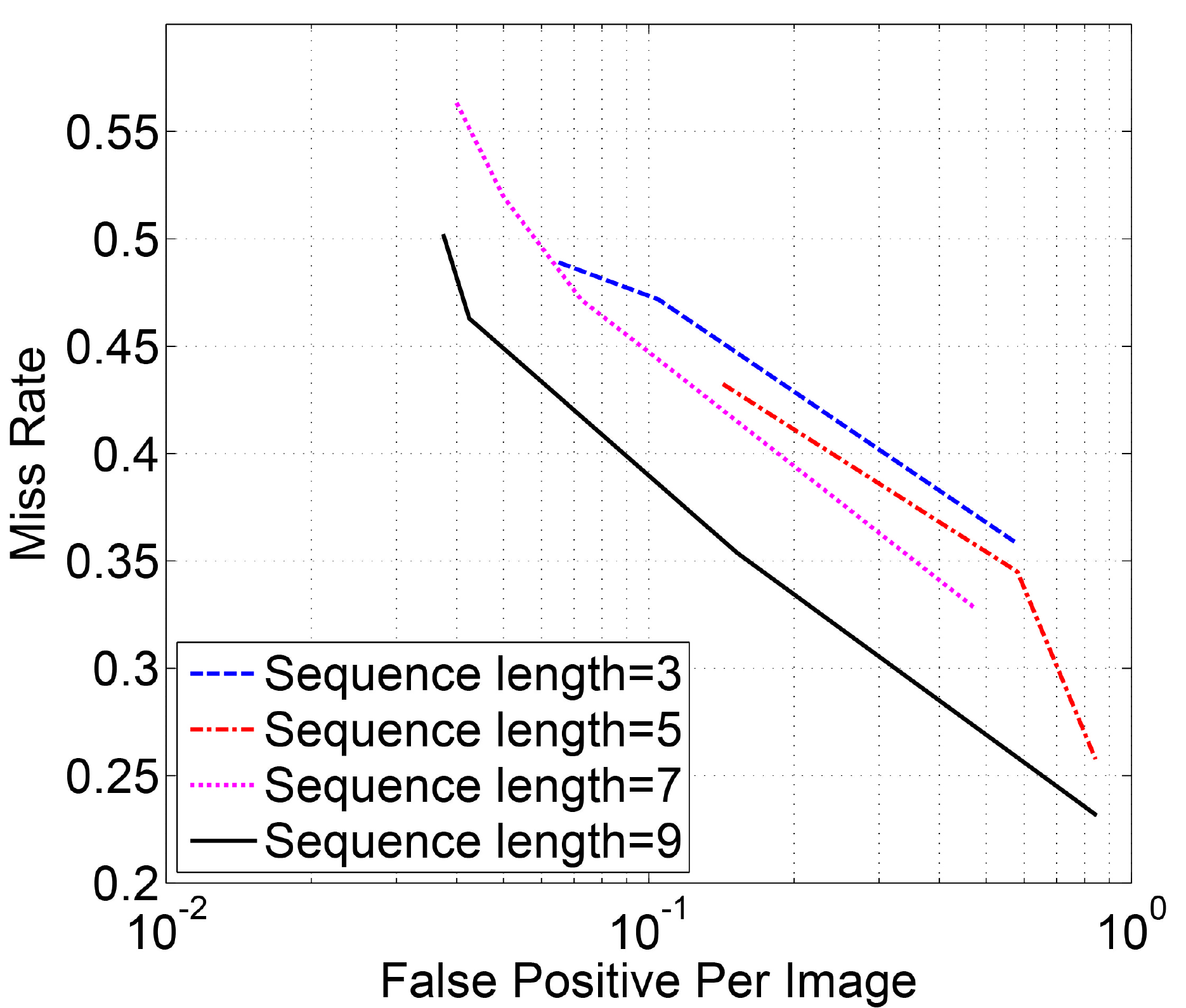}}
%  \vspace{1.5cm}
  \centerline{(c)}\medskip
\end{minipage}
\begin{minipage}[b]{0.50\linewidth}
  \centering
  \centerline{\includegraphics[width=4.15cm]{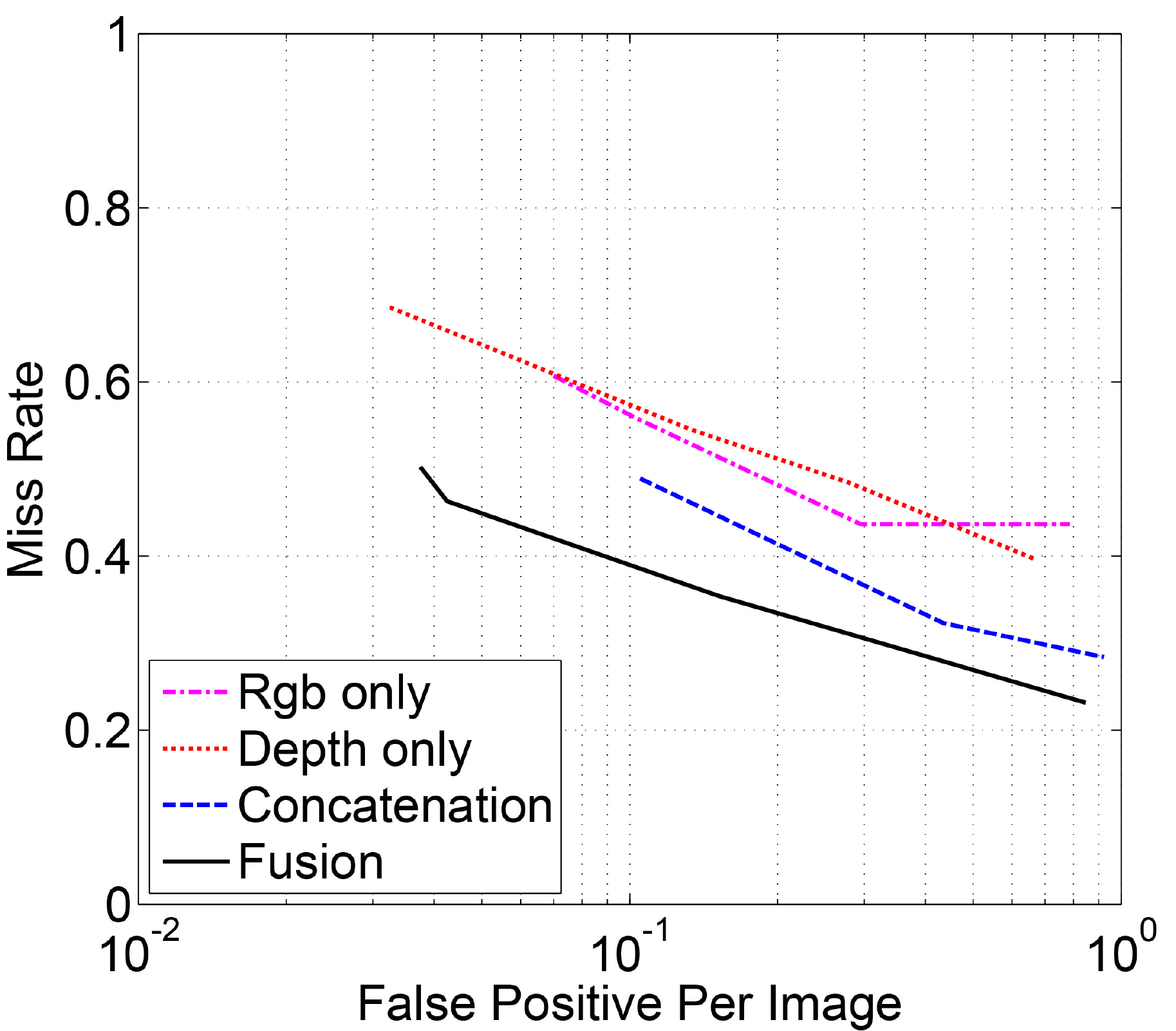}}
%  \vspace{1.5cm}
  \centerline{(d)}\medskip
\end{minipage}
\caption{{\bf Experimental results.} (a) and (b) show results of different methods on \emph{Office Dataset} and \emph{Mobile Dataset}. (c) and (d) demonstrate the comparison of different glimpse numbers and different fusion methods on \emph{Mobile Dataset}.}
\label{fig:results}
\end{figure}

{\bf Datasets.} We evaluate the proposed method on two datasets, which were captured with Kinect at 640 $\times$ 480 resolution.

\begin{enumerate}[(1)]
\item \emph{Kinect Office dataset} \cite{wongui}. This dataset contains 17 video sequences captured in an office. Various human poses such as sitting, standing and walking are included.
\item \emph{Kinect Mobile dataset} \cite{wongui}. This dataset contains 18 video sequences collected by a Kinect mounted on a PR2 robot with a horizontal perspective. The robot was moving inside a building, and challenges like illumination variations and complex background are included. 
\end{enumerate}

{\bf Implementation details.} The generated proposals are labeled binarily as human and non-human. The network learns to predict the binary class with the logistic regression classifier, which takes the output state of the last step of MG-LSTM as input. The objective function takes the negative log-likelihood loss to measure the difference between predicted results and true labels. Back propagation through time (BPTT) is used to minimize the objective function. We set learning rate and decay rate as 0.0004 and 0.97 to train our network. One MG-LSTM layer with the neuron size of 256 is used in our network. In our experiment, for Kinect office dataset, we labeled 1,146 positive samples and 22,835 negative samples for training; for Kinect mobile dataset, we labeled 577 / 23,520 for training. The rest of the frames are used for testing. Since the positive-negative sample ratio is rather low in the training set, in each training epoch, we randomly pick a certain number of negative samples out of the entire negative sample set to handle the unbalanced training samples, which makes the positive-negative sample ratio at approximately 1 to 3 within each epoch.

{\bf Experimental results.} We follow the evaluation protocol in \cite{wongui} and plot the false-positive-per-image (FPPI) vs. miss-rate curves for evaluation.

To evaluate the performance of our proposed method, we compare it with HOG \cite{HOG}, the depth-based detector proposed by Xia et al. \cite{xia}, the RJ-MCMC by Choi et al. \cite{wongui}, the Ring-Wedge Mask method proposed by Liu et al. \cite{jvci} and a deep CNN network proposed by Eitel et al. \cite{depthextract}. Based on the results in Fig. \ref{fig:results}(a) and (b), our method obviously achieves the best performance among all compared methods, which indicates the superiority of our MG-LSTM network with color depth feature fusion. Our method is also proved to outperform deep CNN in this task.

Different sequence length (different glimpse numbers) of MG-LSTM are experimented to assess the effectivity of utilizing multi-scale multi-part information. As shown in Fig. \ref{fig:results}(c), the network with sequence length 9 outperforms others with shorter sequence length, proving that multi-scale information contributes much to the classification performance.

Besides, in order to evaluate the color-depth feature fusion strategy, we compare our network (MG-LSTM with color depth fusion) with three deep neural networks, i.e. MG-LSTM with color only, MG-LSTM with depth only and MG-LSTM with simple color depth concatenation(see Fig. \ref{fig:network1}). In Fig. \ref{fig:results}(d), our proposed color depth feature fusion strategy clearly shows its effectivity since it exceeds the other three information utilization strategy. Generally, experimental results illustrate marked advantage of our method.

\begin{figure}[t]
\begin{minipage}[b]{0.98\linewidth}
  \centerline{\includegraphics[width=8.5cm]{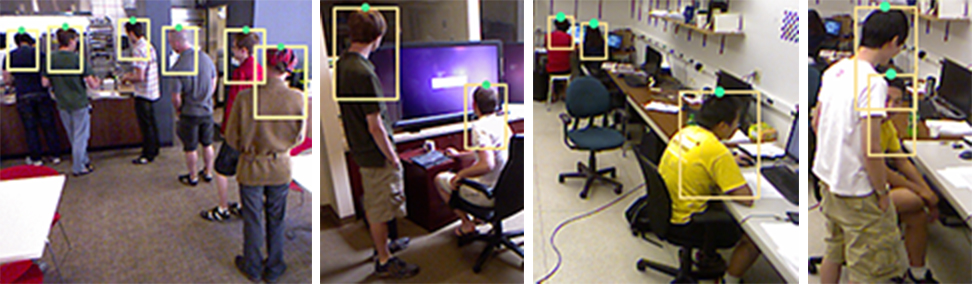}}
\end{minipage}
\caption{{\bf Example of detection results.}}
\label{fig:resultsample}
\end{figure}

\section{Conclusion}
\label{sec:conclusion}

In this paper, we propose a new Multi-Glimpse LSTM network for RGB-D based human detection. In order to better incorporate the RGB and depth information, we further propose a color-depth feature fusion strategy. The comparative experiments show the superior performance of our method on two RGB-D datasets.
\bibliographystyle{IEEEbib}
\bibliography{xudong_li,alexnet,vgg,resnet,ActionRecog1,ActionRecog2,HOD,HZhang,HOG,wongui_choi,cvpr15,HongyangXue,jvci,depthextract,rcnn,segmentation1,segmentation2,objectness,MCG,selectivesearch,xia,final_add1,final_add2,final_add3}

\end{document}